# MRC-GAT: A Meta-Relational Copula-Based Graph Attention Network for Interpretable Multimodal Alzheimer's Disease Diagnosis

Fatemeh Khalvandi, Saadat Izadi and Abdolah Chalechale[*]

Computer Engineering Department, Razi University, Kermanshah, Iran

f.khalvandi@stu.razi.ac.ir, s.izadi@razi.ac.ir, chalechale@razi.ac.ir

## Abstract

Alzheimer's disease (AD) is a progressive neurodegenerative condition necessitating early and precise diagnosis to provide prompt clinical management. Given the paramount importance of early diagnosis, recent studies have increasingly focused on computer-aided diagnostic models to enhance precision and reliability. However, most graph-based approaches still rely on fixed structural designs, which restrict their flexibility and limit generalization across heterogeneous patient data. To overcome these limitations, the Meta-Relational Copula-Based Graph Attention Network (MRC-GAT) is proposed as an efficient multimodal model for AD classification tasks. The proposed architecture, copula-based similarity alignment, relational attention, and node fusion are integrated as the core components of episodic meta-learning, such that the multimodal features, including risk factors (RF), Cognitive test scores, and MRI attributes, are first aligned via a copula-based transformation in a common statistical space and then combined by a multi-relational attention mechanism. According to evaluations performed on the TADPOLE and NACC datasets, the MRC-GAT model achieved accuracies of 96.87% and 92.31%, respectively, demonstrating state-of-the-art performance compared to existing diagnostic models. Finally, the proposed model confirms the robustness and applicability of the proposed method by providing interpretability at various stages of disease diagnosis.

## 1. Introduction

Alzheimer's disease is an irreparable and progressive neurodegenerative disease, mainly affecting older people, causing drastic impairment in cognitive functions, memory loss, and behavioral changes [1]. With the increasing aging population in the entire global health setup, Alzheimer's cases are on the rise, and estimates show that by 2050, more than 130 million people are likely to suffer from Alzheimer's, causing an enormous burden on the entire global health infrastructure. The pre-dementia phase of Alzheimer's, identified as Mild Cognitive Impairment (MCI), is an intermediate phase between aging and dementia, wherein the required cognitive functions are diminished, but the capability to manage day-to-day activities is maintained [2]. The lack of any conclusive treatment [3], underlines the prime importance of early detection and subsequent interventions to control the progression of Alzheimer's and improve the health outcomes [4]. The diagnostic requirements for Alzheimer's generally entailed cognitive function tests like the Mini-Mental State Examination (MMSE) and the Clinical Dementia Rating (CDR) [5], accompanied by neuroimaging techniques, focusing on obtaining structural and functional details on the pathology of Alzheimer's and MCI, describing the abnormalities in the Alzheimer's and MCI-inflicted brains [6]. These techniques, being very cumbersome, require extensive expertise and possess low accuracy rates, thereby underlining the increasing demands and requirements for sophisticated computer-aided diagnostic techniques, possessing enhanced accuracy and efficiency in Alzheimer's diagnostics.

Based on the weaknesses of conventional diagnostic approaches and neuroimaging analysis techniques, various studies have been conducted to develop innovative advanced computing approaches, intending to improve early Alzheimer's disease diagnosis. In early-stage diagnostic studies, conventional machine learning classifiers, including logistic regression, sparse inverse covariance estimation, and multi-kernel SVMs, were used to distinguish between Alzheimer's disease, Mild Cognitive impairment, and Cognitive Normal (CN) and showed moderate accuracy and poor generalization. Later research studies proposed ensemble learning and feature selection techniques, including random forest and RFE-SVM, for improving multimodal feature learning and enhancing diagnostic accuracy [7, 8]. Subsequently, convolutional neural networks (CNNs) showed significant improvements by learning spatial discriminative features from MRI and EEG and transforming into high-resolution 3D and multimodal feature analysis for comprehensive disease modeling and characterization [9-13]. However, those CNN approaches disregard the relational dependencies among different regions of the brain. In this regard, Graph Neural Networks (GNNs) have been identified as robust alternatives to regular CNN approaches, as they can model and capture relational dependences as well as individual variances of the subject-specific data, hence exhibiting superior interpretability and robustness in subject-specific predictive modeling tasks [14-17]. novel advances in multigraph and fusion learning models and approaches have shown improved biomarkers and accuracy by integrating and combining various graphs along with multimodal embedding models [18-20]. In summary, all those latest studies and research show significant and continuous progress toward the paradigmatic shift toward innovative and advanced graphs and multimodal learning approaches with considerations toward leveraging improved accuracy and diagnostic efficiency in Alzheimer's analysis and modeling. Nonetheless, there are still various research challenges and questions yet to be resolved and answered appropriately.

---

[*]Corresponding author

Despite the significant success achieved by the graphical approaches proposed for Alzheimer's disease diagnosis, some issues still need to be resolved to improve generalization and applicability. First, the presence of natural variability between multimodal patterns, including MRI, cognitive, and risk factors, makes it difficult to accurately calculate similarity measures between different individuals, often introducing noisy and volatile graph patterns [18, 19]. In addition, CNN-based models are efficient in learning spatial patterns but inefficient in handling non-imaging features, including demographics and family history, which are essential for fully understanding and modeling the disease [20]. Second, many existing GCNs remain transductive models depending on a fixed graph that restricts their ability to generalize to new, unseen nodes. Hence, they fail to generalize well on novel, unseen subjects and are crippled in conducting standalone tests without rebuilding the entire graphs. Although newly introduced models, such as GAT, and other differentiable components address this drawback, they still depend on large static graphs and sacrifice accuracy [18]. Finally, multimodal attention and fusion techniques, which are useful tools for improving accuracy, are often less interpretable; thus, it becomes difficult to distinguish the influences of contributing and subject-related associations within the entire diagnostic process. In addition, issues of small sample size and varying distributions of features can raise the problem of overfitting. Resolving these issues requires the development of an adaptive, interpretable, and inductive learning model on graphs that can handle multimodal features with varying properties.

To address these challenges, the MRC-GAT model is introduced as a multimodal, episodic meta learning model that integrates copula-based similarity alignment, relational attention, and node-wise gated fusion. Firstly, the features are mapped into a copula feature space through rank Gaussianization, and then robust internal covariances are estimated through Ledoit-Wolf shrinkage within each modality. The pairwise similarities are then measured through Mahalanobis Distance, enabling scale-independent comparison against correlated features. The relation-specific graphs are defined for each modality, and then they are sparsified through directed k-nearest-neighbors (KNN) and a predefined threshold. Meanwhile, for each type of modality, one-hop relational attention and node-wise gated fusion layers are used, and then two-hop attention and secondary fusion refine the subject embedding by extending the receptive field and improving learning stability. Importantly, attention weights $\alpha_{ij}^{(g)}$ and node-wise fusion weights $\gamma_i^{(g,\cdot)}$ are quantifying the contributions of nodes and neighbors, providing transparent and clinically coherent interpretation. Finally, an episodic meta-learning protocol trains the model to classify unseen query subjects from small support sets, ensuring robust generalization and improved learning stability. Experiments conducted on the TADPOLE and NACC datasets show that MRC-GAT delivers strong and consistent diagnostic performance, achieving 96.87% and 92.31% accuracy, and providing stable multimodal and class discrimination across heterogeneous subjects. In summary, the following are the key contributions of this research work:

- To reduce instability in similarity measurement for RF, COG, and MRI, a copula-aligned graphical construction process is proposed.

- A two-stage relational attention modeling with node-wise gated fusion is proposed, which enables explanation of attention weights on each edge and each modality.
- An episodic meta-learning approach is proposed to improve the generalization capability of the model, and it can make inductive inference on novel instances and achieve stable performance on various data sets.

The remainder of this article is structured as follows. In Section ۲, related work is reviewed; Section ۳ describes the methodology and details of the proposed architecture; Section 4 outlines the evaluation setup, datasets, evaluation metrics, and reports experimental results; and Section 5 concludes the paper by summarizing the key findings and suggesting potential directions for future research.

## 2. Related Work

In this section, the current progress in multimodal fusion, graph learning, and Explainable AI in Alzheimer's disease diagnosis will be examined. The section will begin with the discussions on multimodal fusion techniques, followed by an analysis of graph-based approaches, and finally describe emerging techniques improving the interpretability and accuracy of Alzheimer's disease diagnostic modeling.

### 2.1 Multimodal Fusion Approaches

The multimodal fusion techniques used in diagnosing Alzheimer's disease focus on combining diverse clinical, cognitive, neuroimaging, and genetic features into a unified modeling scheme. More recently, studies on multimodal fusion in Alzheimer's disease can be primarily clustered into two overarching paradigms, namely feature-level fusion and decision-level fusion. At feature level fusion, cascaded deep learning models, as exemplified by the multimodal mixing transformer (3MT), processed clinical and neuroimaging features together through the mechanisms of cross-attention and modality dropout, obtaining robust generalization under missing-modality scenarios [21]. Similarly, another deep learning architecture, the deep multimodal discriminative and interpretability network (DMDIN), achieved feature alignments through the use of multilayer perceptrons and generalized canonical correlation analysis, creating a discriminative shared space with improved separation and identification of distinct patterns in the various modalities [22]. In another line of work, interpretable models of disease progression have utilized MRI, clinical scoring, and genetic polymorphisms through interaction models, boosting robustness against center variability [23]. Meanwhile, other trimodal fusion techniques feature successful discrimination between progressive and stable MCI through the combined use of SNPs, gray-matter ratios, and sMRI features, underscoring the morphological predominance of gray-matter ratios [24]. At the decision level, various late-fusion methods combine the predictions of modality-specific learners. For example, a multi-level stacking ensemble fuses six base classifiers per modality and combines the predictions at a later stage across modalities, leading to increased accuracy and enhanced interpretability due to optimized feature selection [25]. Multimodal mixing methods couple MRI-tailored vision transformers with 1D-CNNs, utilizing multi-scale attention for clinical features, further boosting the performance in AD recognition tasks [26]. The presented collection of multimodal fusion studies underlines the importance of bringing together complementary biomarkers in addition to identifying challenging aspects of finding appropriate nonlinear cross-modal relationships and maintaining interpretability when complex architectures are being modeled for fusion.

### 2.2 Graph-Based Methods for AD Diagnosis

Graph convolutional formulations have been used to encode subject-to-subject relationships from imaging-derived biomarkers and phenotypes. The UNB-GCN method builds edges from phenotypic similarity and uses attention to refine morphological biomarkers of cortical atrophy. Experiments on ADNI show improved accuracies for AD vs. CN and AD vs. MCI, while the authors report that UNB better captures AD-induced cortical changes compared to traditional volumetrics [27]. Beyond fixed graphs, an auto-metric GNN introduces a metric-based meta-learning strategy that trains on many small node-classification tasks to enable inductive testing; an AMGNN layer with a probability constraint learns node-similarity metrics while fusing multimodal data, yielding high accuracy on TADPOLE for both early diagnosis and MCI to AD conversion [18]. Furthermore, a multigraph-combination screening GCN constructs numerous graphs spanning multiscale features and multi-hop neighborhoods and then selects optimal combinations via a learned predictor, followed by multigraph attention; this design mitigates incorrect edges and improves robustness on NACC and TADPOLE [20]. Besides, a feature-aware multimodal method integrates

SHAP-based boosting feature selection, cross-modal attention to model subtle inter-modality relations, and a GCN branch for heterogeneous data, with an automatic model-fusion strategy learning weights between sub-models; results on two ADNI cohorts indicate strong diagnostic performance [28]. On the other hand, an interactive deep cascade spectral GCN constructs separate imaging and non-imaging relational graphs with learnable edge generators, and employs dual cascade spectral branches with inter-branch interaction to capture complementary semantics across depths, surpassing prior state of the art on multiple disease datasets [29].

## 2.3. Explainable and Trustworthy AI for AD Diagnosis

Explainability for the diagnosis of AD has been investigated through various paradigms. The deep multi-modal discriminative and interpretability networks were used to realign the various modalities on a discriminative subspace, and through knowledge distillation, they were able to revive synchronized representations, stressing crucial ROI's involved in the classification decision of AD [22]. The case-based method for counterfactual reasoning combined U-Net and GAN models to provide subject-specific maps, explaining how slight morphological modifications might affect the diagnostic decision, without affecting the excellent accuracy on tasks [30]. Also, attention-enhanced autoencoders combined with Grad-CAM aided in locating those parts of the brain crucial for decision-making within T2-weighted sMRI inputs, leading to high accuracy and useful saliency maps [31]. In addition, multimodal models and the federated learning platform integrating Random Forest with SHAP explanation unraveled feature attributions through MRI segmentation, clinical, and psychological input data while retaining privacy and achieving strong precision, recall, and Area under Curve (AUC) [32]. To overcome the limitations identified in some current graph-based models and multimodal classification approaches, the MRC-GAT model has been developed to provide an improved, more adaptive, and dependable method for Alzheimer's disease classification. Unlike other graph-based models, which relied on fixed graph topology [20, 27, 29], or models based on individual auto-metric relationships for meta-tasks [18], MRC-GAT has an inductive architecture especially designed to address the above challenges and provide a reliable integration of heterogeneous multimodal data. A concise comparison with other methods is provided in Table 1.

Table 1. Summary of related works.

| Reference | Year | Method | Dataset | Advantage | Limitation |
|---|---|---|---|---|---|
| [18] | 2021 | Auto-Metric GNN with meta-learning | TADPOLE | Enables inductive testing and learns adaptive node-similarity metrics | Requires multiple small graph tasks; limited interpretability |
| [20] | 2024 | Multigraph screening + multigraph attention | TADPOLE, NACC | Reduces noisy edges and enhances node aggregation among similar patients | High computational cost due to graph enumeration; Limited scalability to large cohorts |
| [21] | 2023 | Cross-attention + modality dropout | ADNI | Robust to missing modalities | Requires careful tuning of transformer layers; May overfit on small complete subsets |
| [22] | 2023 | MLP + GCCA + knowledge distillation | ADNI | Discriminative embeddings; Identifies significant ROIs via distillation | Complex optimization process; Sensitive to hyperparameters |
| [23] | 2024 | Deep model with interaction encoding | ADNI | Improves long-term prediction accuracy; Integrates interaction effects via multimodal data | Requires large multi-center data; Sensitive to scanner/inter-center variability |
| [24] | 2025 | Trimodal fusion (SNP + RGV + sMRI) | ADNI | High accuracy distinguishing sMCI vs pMCI; RGV plays key role in morphological discrimination | Limited interpretability; Requires complete trimodal data |
| [25] | 2023 | Stacking ensemble + PSO-based feature selection | ADNI | Multi-modality stacking; improves accuracy and interpretability | Depends on handcrafted sub-score; Complex 3-level stacking increases training time |
| [26] | 2025 | MRI_ViT + 1D-CNN with attention fusion | ADNI | Robust multimodal fusion; Leverages ViT for spatial and 1D-CNN for tabular features; Strong accuracy | Requires large labeled datasets; High computational demand |
| [27] | 2023 | UNB-GCN with attention | ADNI | Highlights key cortical regions; interpretable | Single-modality focus; limited generalization |
| [28] | 2024 | SHAP-based selection + GCN + auto-fusion | TADPOLE | Preserves low-dim correlations via SHAP boosting; Captures subtle cross-modal relations | Complex integration; risk of overfitting |
| [29] | 2024 | Dual spectral GCN branches | ADNI, ABIDE | Builds multi-relational graphs from imaging & non-imaging; Deep cascade interaction enriches high-level features | High computational complexity; Requires learnable edge generators per modality |
| [30] | 2025 | U-Net + GAN counterfactuals | ADNI | Generates causal counterfactual maps; Outperforms SOTA in ACC and AUC; Outperforms Grad-CAM & other XAI methods | High model complexity; computationally heavy |
| [31] | 2025 | Autoencoder + Grad-CAM | ADNI | Highlights discriminative brain regions | Single-modality (sMRI only); Limited to 2D axial slices |
| [32] | 2025 | Federated RF + SHAP explainability | OASIS | Preserves data privacy and interpretability | Limited to classical models; No joint feature learning across institutions |
| **Proposed Work** | | | TADPOLE, NACC | - Aligns data distributions across different modalities.<br>- Provides interpretable attention weights across modalities.<br>- Enables inductive generalization to unseen subjects through episodic meta-learning | - Requires tuning of k and copula parameters.<br>- Computationally heavier than single-relational GCNs. |

## 3. Methodology

This section describes the MRC-GAT method, Section 3.1 provides problem statement and explains the motivation for the proposed model. Section 3.2 formalizes multimodal classification under an episodic meta-learning setting and introduces the notation. Section 3.3 describes the model architecture. Finally, Section 3.4 outlines the training strategy.

### 3.1. Problem Statement

Alzheimer's disease represents a progressive form of neurodegeneration associated with cognitive decline and brain shrinkage, and it's precisely this early stage in Alzheimer's disease, presented as mild cognitive impairment (MCI), which holds great significance for early diagnosis. Despite large amounts of research involving developments in machine learning and imaging techniques, early diagnosis constitutes a difficult task due to variability in clinical data originating from several sources, such as demographics, cognitive evaluation, and MRI scan. The graph diagnosis approaches have several challenges in this context. The heterogeneity of data makes it difficult to build similarity graphs across subjects. The standard GNN models have static graph structures, which limit their ability to generalize to unseen subjects. Moreover, many existing fusion approaches provide only limited transparency, which reduces their practicality in clinical settings. In view of these issues, it becomes necessary for an ideal diagnosis model to be able to align data from diverse modalities systematically, and be able to generalize to new subjects, as well as provide for the interpretability of results. To fulfill these requirements simultaneously, it becomes necessary to combine copula-based techniques, episodic meta-learning approaches, and relational attention methods via node-wise gated fusion to form the MRC-GAT model.

## 3.2. Problem Formulation

Let $D = \{(x_i, y_i)\}_{i=1}^{N}$ be the dataset, where $x_i \in \mathbb{R}^F$ is the multimodal feature vector of the subject $i$, and $y_i \in \{1, \ldots, C\}$ is the diagnostic label (CN, MCI, AD). Each subject vector is partitioned into three modalities:

$$x_i = [\, x_i^{RF},\ x_i^{COG},\ x_i^{MRI}\,] \tag{1}$$

where $x_i^{RF}$ corresponds to demographic risk factors, $x_i^{COG}$ is cognitive test scores, and $x_i^{MRI}$ denotes MRI features. The objective is to learn a function:

$$f: \mathbb{R}^F \to \{1, \ldots, C\} \tag{2}$$

such that the predicted label $\hat{y}_i = f(x_i)$ is close to the true label $y_i$. with this in mind, the dataset has been represented as a multi-relational graph:

$$\mathcal{G} = (V, \{E^{(g)}\}_{g \in \{RF, COG, MRI\}}) \tag{3}$$

where $V$ contains the nodes representing the subjects, and each relation type $E^{(g)}$ embodies all pairwise dependencies defined within one modality. The learning objective is thus set as minimizing the following classification loss over the episodic tasks:

$$\min_{\theta} \mathbb{E}_{\mathcal{S} \sim \mathcal{D}}[\, \mathcal{L}(f_\theta(\mathcal{G}_\mathcal{S}), Y_\mathcal{S})\,] \tag{4}$$

where $\theta$ are the model parameters, $\mathcal{S}$ is a sampled meta-task including support and query nodes, and $\mathcal{G}_\mathcal{S}$ denotes the corresponding multi-relational graph.

## 3.3. Model architecture

The entire training process of the proposed MRC-GAT is shown in Fig. 1, which describes the meta-learning process over various episodes. In the training process, for each training iteration, samples of episodes are collected from the training set, and each episode is a small graph generated by well-balanced support set samples and an unlabeled query node. The graphs are input into the model simultaneously, and the forward and backward processes calculate the loss for the episode, and the model updates the parameters once for each training iteration. Then, the trained parameters will be transferred to the next training iteration, by which the model continually trains to acquire the relational representations. After training iterations, the trained parameters are directly used for inference on other episodes without any parameter fine-tuning. The pseudo-code of the proposed main model is shown in Algorithm 1, and Fig. 2 describes the exact process and details of the MRC-GAT model. The remaining content of this section is divided into the following parts:

- Episodic Task Design and Batch Construction;
- Copula-Based Multi-Modal Similarity Computation;
- Graph Construction and Sparsification;
- One-hop Relational Graph Attention;
- Node-Wise Gated Fusion Across Modalities;
- Two-hop Relational Graph Attention and Fusion;

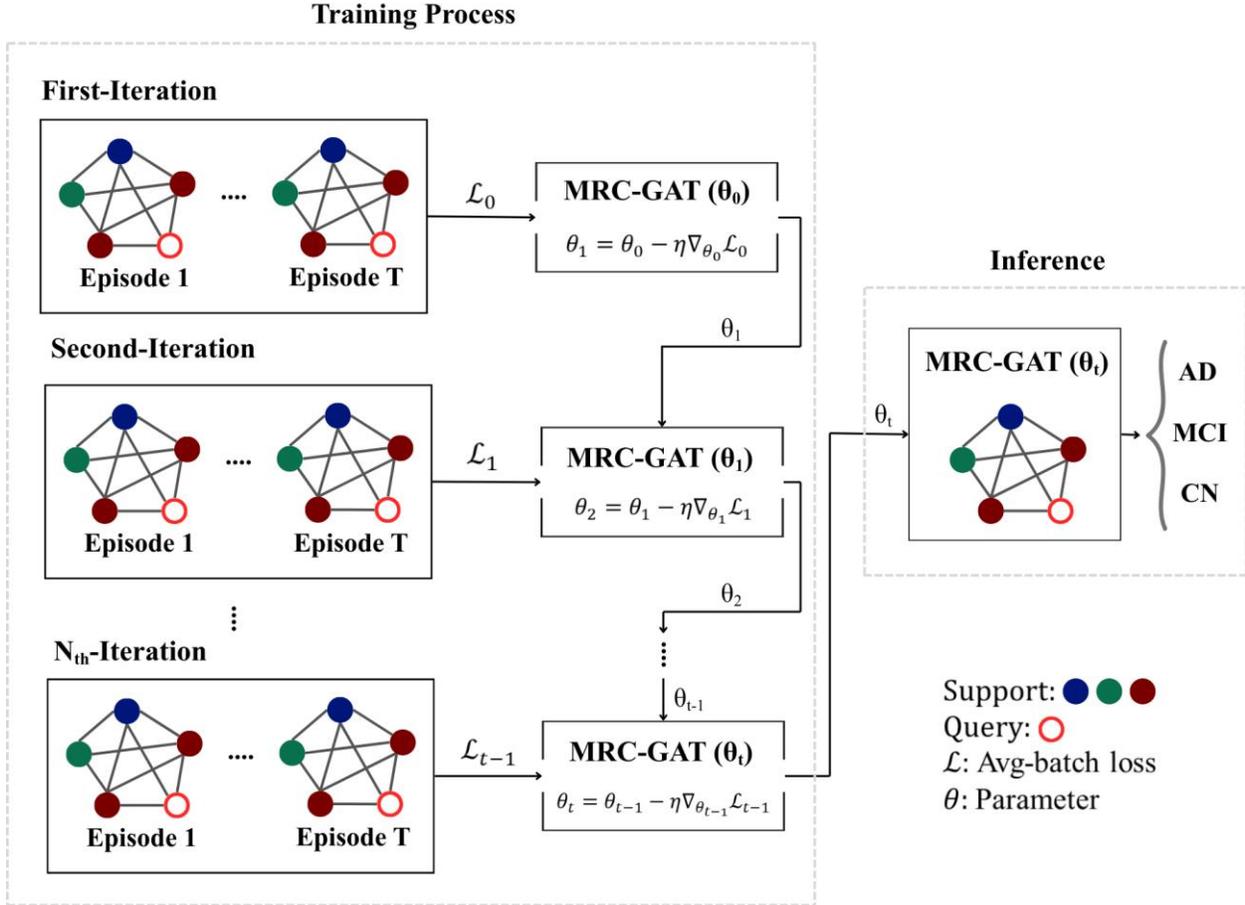

**Fig. 1.** Overview of episodic meta-training/ inference for MRC-GAT. Multiple episodes, consisting of support and query samples, are processed in each iteration, with the average loss being computed and the model parameters subsequently updated. Finally, the trained model performs inductive inference to classify new query subjects into CN, MCI, or AD groups.

### 3.3.1. Episodic Task Design and Batch Construction

The Alzheimer's disease diagnosis is formulated as the following supervised node classification problem within an episodic meta-learning method. Each episode $\mathcal{T}$ is defined as a small, encapsulated classification task consisting of a support set $S$ and a query node $Q$, and is mathematically expressed as:

$$S = \bigcup_{c=1}^{C} \{(x_i, y_i = c)\}_{i=1}^{q}, \qquad Q = \{\hat{x}\} \qquad (5)$$

where $C$ is the total number of diagnostic classes, $q$ is the number of support samples to be chosen from each class. $x_i \in \mathbb{R}^F$ is the multimodal feature vector of subject $i$. $y_i$ is the ground-truth diagnostic label and $\{\hat{x}\}$ is the query node, which is to be classified and is labeled as unidentified. Hence, the total number of nodes in this episode will be $N = C \times q + 1$. The rationale behind this setting is to provide input to the model in an equal manner for all classes, along with one unidentified query node per episode, which is required to be inductively classified by the model. Additionally, within each episode, the union of the support and query nodes, denoted as $S \cup Q$, constitutes the set of vertices for three relation-specific graphs for each modality, namely $\mathcal{G}_{RF}$, $\mathcal{G}_{COG}$, $\mathcal{G}_{MRI}$. Every relation-specific graph reflects the inherent patterns of similarities for its respective modality, yet these distributions of raw features have extreme variations from each other in different modalities. To facilitate similarities to be computed, there is a need for these multimodal features to be projected into an aligned common space via the transformation based on copulas, in which similarities in terms of distances/edge weights in different modalities become comparable.

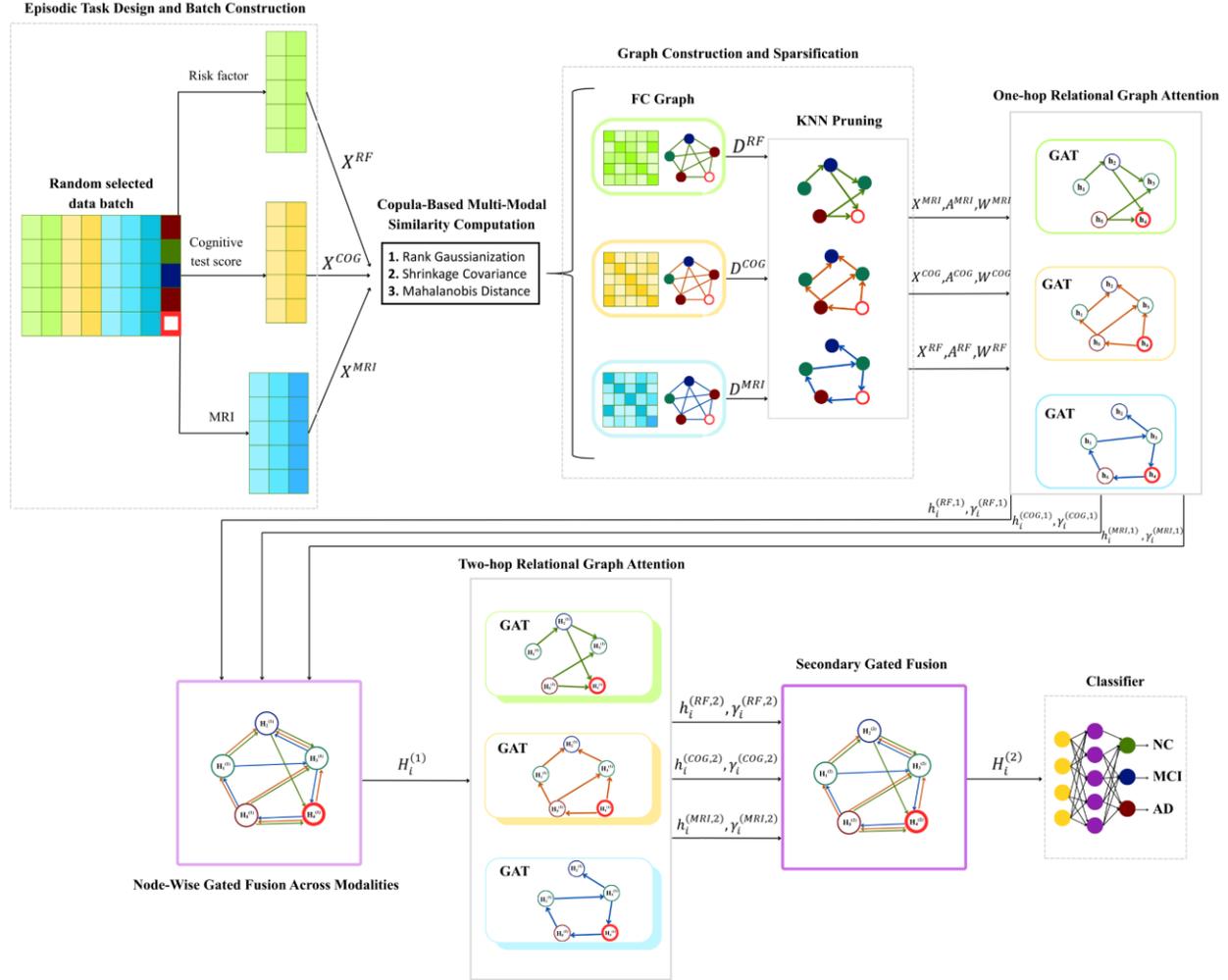

**Fig. 2.** Model architecture and workflow of the MRC-GAT method.

### 3.3.2. Copula-Based Multi-Modal Similarity Computation

Following the episodic task design outlined above, this section is devoted to calculating pairwise similarities between risk factors, cognitive data, and MRI modalities. To achieve statistical comparability across heterogeneous feature spaces, raw feature vectors are mapped into a Gaussian copula domain, which enables more accurate modeling of nonlinear dependencies and covariance structures. The overall procedure is organized in three steps: rank Gaussianization of feature distributions, covariance shrinkage to stabilize dependency estimates, and Mahalanobis distance computation to measure the statistical dissimilarity between subjects. These relationships are computed in three steps, as discussed below:

- **Rank Gaussianization:** To normalize heterogeneous features across modalities, each feature dimension is first transformed into the Gaussian copula domain as:

$$Z = \Phi^{-1}\left(\frac{r+1}{N+1}\right) \tag{6}$$

where $r$ is the rank of a value among $N$ samples, and $\Phi^{-1}(\cdot)$ is the inverse standard normal cumulative distribution function (CDF). This transformation enforces marginal Gaussianity so that making features from different modalities are statistically comparable while preserving their rank-based dependencies can still be preserved.

- **Covariance Shrinkage:** The Covariance patterns in relation $g$ are regularized by Covariance shrinkage named Ledoit-Wolf shrinkage, defined as:

$$\Sigma_g = (1-\lambda)\frac{Z_g^\top Z_g}{N-1} + \lambda \cdot \frac{tr\left(\frac{Z_g^\top Z_g}{N-1}\right)}{d_g} I \tag{7}$$

where $Z_g$ is the Gaussianized feature matrix for the relation $g$, $d_g$ is the number of features in relation $g$, $I_{d_g}$ is the identity matrix of size $d_g \times d_g$, while $tr(\cdot)$ is the matrix trace operator, and $\lambda \epsilon [0,1]$ is the shrinkage parameter. The use of the shrinkage parameter here reduces the instability of the estimated value, which may be problematic in high-dimensional data and small samples.

- **Mahalanobis Distance Between Subjects:** The Mahalanobis distances are calculated between all pairs of subjects for relation $g$ as follows:

$$D_{ij}^{(g)} = (Z_{i,g} - Z_{j,g})^\top \Sigma_g^{-1} (Z_{i,g} - Z_{j,g}) \tag{8}$$

where $Z_{i,g}$ and $Z_{j,g}$ are the Gaussian feature vectors of subjects $i$ and $j$ under relation $g$, $\Sigma_g^{-1}$ is the inverse of the regularized covariance matrix, and $D_{ij}^{(g)}$ is the Mahalanobis distance between subjects $i$ and $j$ in relation $g$, defined as the quantification of the statistical dissimilarity between the two subjects considering the correlation structure between features. The inclusion of covariance information in this measure implies that this measure calculates distances within the whitened feature space, assigning more significance to correlated features. The set of all pairwise Mahalanobis distances constitutes the statistical basis for generating relation-specific graphs, within which each node pair is connected based on their similarity measurements.

### 3.3.3. Graph Construction and Sparsification

Using the pre-computed pairwise Mahalanobis distances, an initial fully connected graph is created for each relation, with edge weights defined as inverse functions of the respective distances. The adjacency and weight matrices are defined as:

$$A_{ij}^{(g)} = \begin{cases} 1, & i \neq j \\ 0, & \text{otherwise} \end{cases}, \quad w_{ij}^{(g)} = \frac{1}{1 + D_{ij}^{(g)}} \tag{9}$$

here, $A_{ij}^{(g)}$ is a binary adjacency matrix indicating whether there is an edge between nodes $i$ and $j$, and the similarity weight $w_{ij}^{(g)}$ is inversely proportional to the Mahalanobis distance $D_{ij}^{(g)}$. Thus, a subject with a higher similarity value, i.e., smaller distance, will be more strongly weighted in this process. At this point, the graph is fully connected and dense. Upon this graph initialization, a KNN-based sparsification step is applied to retain only the most relevant local relationships and suppress weak or non-informative connections. This pruning strategy facilitates communication between any node, primarily with its most relevant neighbors, while maintaining information about the structure. These k-nearest neighbors for every node $i$, given relation $g$ is determined by:

$$N_k^{(g)}(i) = arg\ topk_{j \neq i}\left(-D_{ij}^{(g)}, k\right) \tag{10}$$

In this case, $N_k^{(g)}(i)$ represents the index set of the $k$ nodes most similar to node $i$. The sparse matrices are then expressed as follows:

$$\tilde{A}_{ij}^{(g)} = \begin{cases} 1, & j \in N_k^{(g)}(i) \\ 0, & \text{otherwise} \end{cases}, \qquad \widetilde{W}_{ij}^{(g)} = \tilde{A}_{ij}^{(g)} \cdot \frac{1}{1 + D_{ij}^{(g)}} \tag{11}$$

where $\tilde{A}_{ij}^{(g)}$ is defined as whether node $j$ is within the top-$k$ neighbors of node $i$ under relation $g$, and $\widetilde{W}_{ij}^{(g)}$ assigns weights based on the inverse of the Mahalanobis distance. The value of the hyperparameter $k$, therefore, directly influences the balance between sparsity and coverage of information within graphs. To further refine the structure, the distance gating mechanism is used in filtering out the low or noisy edges with similarity less than the predefined value $\tau$. The process is defined as:

$$M_{ij}^{(g)} = 1\left(D_{ij}^{(g)} \le \tau\right), \qquad \widehat{W}^{(g)} = \widetilde{W}^{(g)} \odot M^{(g)} \tag{12}$$

In this context, $\tau > 0$ is the maximum allowed value for the retention of an edge, and $\odot$ is element-wise multiplication, with the binary mask $M^{(g)}$ eliminating edges above this value. The above process serves as the quality control process, whereby only strong and trusted edges are retained. Moreover, the obtained graph is considered directed, implying that the node $j$ may be among the top $k$ neighbors of the node $i$ without being mutually connected. This is more flexible than depicting dependencies among subjects, as it models the complex directed relationships typically recurring within clinical or biological networks. Consequently, this directed and sparse-graph structure constitutes a basis for the following relation-specific attention mechanism, used for adaptive information aggregation through modality-dependent edges.

### 3.3.4. One-hop Relational Graph Attention

In this section, the relational graph attention mechanism is introduced on top of the sparsified graph structure developed earlier. During this stage, the model leverages the directed neighborhood structure to learn modality-specific node representations through the use of attention-driven message passing. Unlike conventional undirected settings, here the directionality of edges makes it possible for the model to selectively aggregate from incoming neighbors, capturing asymmetric dependencies. This step forms the foundation for learning local relational patterns before integrating information across modalities. The input feature vector at the current layer is represented by $h_i$ for node $i$ in relation $g$ denote its input feature vector at the current layer. At layer one, this will refer to the input multimodal features ($h_i = x_i$), whereas at other layers, it will be represented by the fused feature obtained through the previous step ($h_i = H_i^{(1)}$). To capture the directed relationships between the nodes, an attention coefficient is calculated for the directed edge $j \to i$ as follows:

$$\alpha_{ij}^{(g)} = softmax_j\left(LeakyReLU\left((a_g)^\top [W^{(g)} h_i \parallel W^{(g)} h_j]\right)\right) \tag{13}$$

where $W^{(g)} \in \mathbb{R}^{d' \times d}$ is a relation-specific projection matrix mapping node features into a latent subspace of dimension $d'$, $a_g \in \mathbb{R}^{2d'}$ is a trainable attention vector that captures pairwise feature interactions, $[\cdot \parallel \cdot]$ denotes vector concatenation. After computing the attention weights, each node $i$ updates its representation in relation $g$ as follows:

$$h_i^{(g,1)} = \sigma\left(\sum_{j \in N^{(g)}(i)} \alpha_{ij}^{(g,1)} W^{(g,1)} h_j\right) \tag{14}$$

where $h_i^{(g)}$ is the updated node embedding for the relation $g$, $N^{(g)}(i)$ denotes the set of neighbors connected to $i$ in the $g$-specific graph, $\alpha_{ij}^{(g)}$ represents the normalized attention coefficients, and $\sigma(\cdot)$ is the Exponential Linear Unit (ELU) activation, which stabilizes gradients and introduces nonlinearity. By doing so, messages from feature-similar neighbors are selectively aggregated to yield richer relation-specific embeddings. This node update procedure enriches each subject's representation by selectively aggregating information from its most similar feature neighbors within each modality. Based on the enriched embeddings learned above, the subsequent step develops a node-wise gated fusion approach to adaptively integrate information across risk factor, cognitive, and MRI modalities to represent each node.

### 3.3.5. Node-Wise Gated Fusion Across Modalities

After obtaining relation-wise embeddings $h_i^{(g)}$, the model combines them by using learning-enabled gating components, which control the contribution of various modalities toward the fused node representation. The combination is expressed as:

$$s_i^{(g,1)} = \left(u_1^{(g)}\right)^\top h_i^{(g,1)} \tag{15 - a}$$

$$\gamma_i^{(g,1)} = \frac{exp\left(s_i^{(g,1)}\right)}{\sum_{g'} exp\left(s_i^{(g',1)}\right)} \tag{15 - b}$$

$$H_i^{(1)} = \sum_{g \in \{RF, COG, MRI\}} \gamma_i^{(g,1)} h_i^{(g,1)} \tag{15 - c}$$

where $u_g \in \mathbb{R}^{d'}$ is a learnable gating vector for relation $g$, $s_i^{(g,1)}$ is the unnormalized gate score, which calculates the weightage of modality $g$ for node $i$, $\gamma_i^{(g,1)}$ is the normalized gating coefficient obtained by the softmax operation on all the modalities; and $H_i^{(1)}$ is the fused embedding in the first layer, which aggregates all the relational representations into one embedding. This adaptive gating procedure enables the network to emphasize informative modalities and attenuate less relevant ones on a node-specific basis. Building on top of these fused embeddings, the subsequent two-hop relational attention layer extends contextual reasoning and captures higher-order dependencies across directed, modality-specific relational graphs.

### 3.3.6. Two-hop Relational Graph Attention & Fusion

The embeddings $H_i^{(1)}$ from the first layer are then broadcast to all three modality-specific graphs and input into the second relational GAT layer. This layer allows two-hop reasoning and computes node representations by gathering information from the node's neighbors and the neighbors' neighbors in the directed relation, expressed as follows for the node $i$ and relation $g$:

$$h_i^{(g,2)} = \sigma\left(\sum_{j \in N^{(g)}(i)} \alpha_{ij}^{(g,2)} W^{(g,2)} H_j^{(1)}\right) \tag{16}$$

where the second-layer projection matrix is denoted by $W^{(g,2)} \in \mathbb{R}^{d' \times d''}$, and attention coefficients are represented as $\alpha_{ij}^{(g,2)}$, while the input embedding obtained from the first fusion level is $H_j^{(1)}$. A new gating mechanism is then applied to fuse the relation-specific embeddings at this deeper level; this mechanism preserves the same mathematical structure as Eqs. 16 (a) – (c), but it now operates on the updated representations $h_i^{(g,2)}$, which captures much more contextual information extracted by multi-hop propagation. In this case, the learnable gate for the relation $g$ is denoted $u_2^{(g)}$, while the corresponding coefficient $\gamma_i^{(g,2)}$ is the normalized gating weight for modality $g$, and the outcome $H_i^{(2)}$ is the final fused embedding, which captures multimodal relationships across modalities.

The second relational-GAT layer increases the receptive field from one-hop to two-hop neighborhoods, allowing for information propagation along both intra-modal and cross-modal connections. This hierarchical attention structure develops a broader contextual integration and mitigates the over-smoothing impact of deeper GNNs [33, 34]. Modality contributions adaptively fuse in a node-wise gated manner, maintaining subject-level differences and avoiding feature homogenization across layers. As a result, final embeddings $H_i^{(2)} = \{H_i^{(2)}\}_{i=1}^{N}$ capture both fine-grained local affinities and global relational context spanning clinical, cognitive, and neuroimaging domains. These enriched representations then serve as input to the downstream classifier, where diagnostic predictions for query nodes are generated in episodic training.

### 3.4. Episodic Meta-Learning Strategy

In this section, the training process is arranged in an episodic meta-learning loop as illustrated in Fig. 1. At iteration $t$, a batch of $B$ episodes $\{\mathcal{T}_t^{(b)}\}_{b=1}^{B}$ is sampled; each episode contains a balanced support set and one unlabeled query node. These samples are passed through the MRC-GAT stack using the current parameters $\theta_{t-1}$. For each episode $\mathcal{T}_t^{(b)}$, the model calculates the prediction for the query node and determines the $\ell_t^{(b)}$. The iteration-level objective is then defined as the average loss over all $B$ episodes:

$$\mathcal{L}_t = \frac{1}{B}\sum_{b=1}^{B} \ell_t^{(b)} \tag{17}$$

where $B$ denotes the number of episodes per iteration, $\ell_t^{(b)}$ is the focal loss for the episode $b$, and $\mathcal{L}_t$ denotes the batch-averaged episodic loss used for the update. Then, a single meta-update is applied to model parameters as follows:

$$\theta_t = \theta_{t-1} - \eta \nabla_{\theta_{t-1}} \mathcal{L}_t \tag{18}$$

where $\theta_{t-1}$ is the model parameter before the update, $\eta$ is the learning rate, and $\nabla_{\theta_{t-1}} \mathcal{L}_t$ is the gradient of the averaged episodic loss. This update scheme, in which one update is performed per iteration, ensures that each gradient step reflects the diversity of multiple independent tasks, rather than a single episode, promoting stable and generalizable learning. The final parameters $\theta_t$ after $\mathcal{N}$ iterations are used directly for inference on unseen episodes without further fine-tuning, as shown in the right panel of Fig. 1. Given a new support-query configuration, it performs relational reasoning and gated fusion exactly as during training, while preserving the episodic structure. This allows the model to perform inference on subjects never seen before. Message passing is performed via relational GAT layers, where information from various modalities is adaptively integrated using node-wise gating, which provides the fused representation $H_i^{(2)}$ for node $i$. Then, the query node embedding is passed through a two-layer multilayer perceptron classifier (MLP) to estimate the diagnostic probability distribution:

$$\hat{\mathcal{Y}}_i = Softmax(W_2 ReLU(W_1 H_i^{(2)} + b_1) + b_2) \tag{19}$$

where $W_1$ and $W_2$ are learnable weight matrices, $b_1$ and $b_2$ are bias terms, and $ReLU(.)$ denotes the rectified linear activation function, and the output $\hat{\mathcal{Y}}_i \in \mathbb{R}^C$ represents a probability vector over the $C$ diagnostic categories. Finally, the training objective can be written as:

$$\min_{\theta} \mathbb{E}_{\mathcal{S} \sim \mathcal{D}}[\mathcal{L}_{focal}(f_\theta(\mathcal{G}_\mathcal{S}), Y_\mathcal{S})] \tag{20}$$

where $\theta$ denotes all trainable parameters, $\mathcal{S}$ is a sampled episodic task drawn from the distribution $D$, $f_\theta$ denotes the MRC-GAT model parameterized by $\theta$, and $\mathcal{L}_{focal}$ represents the focal loss that balances easy and hard examples in addition to tackling class imbalance. This meta-learning formulation allows the model to learn representations inductively across heterogeneous graphs, which are generalizable to unseen query subjects.

---

**Algorithm 1. Training of MRC-GAT with Episodic Meta-Learning**

Input: training set $s$ with label {CN, MCI, AD}
Output: Trained parameters $\theta$

1. for each meta-epoch $t = 1 \ldots T$ do
2.    Sample support nodes and one query node per class;
3.    Construct relations {RF, COG, MRI};
4.    for each relation g do
5.      $Z_g = \Phi^{-1}((r+1)/(N+1))$     # Gaussian copula normalization
6.      $\Sigma_g \leftarrow = (1-\lambda)\frac{Z_g^T Z_g}{N-1} + \lambda \cdot \frac{tr\left(\frac{Z_g^T Z_g}{N-1}\right)}{d_g} I$     # Ledoit–Wolf covariance estimation
7.      $D_{ij}^{(g)} \leftarrow (Z_{i,g} - Z_{j,g})^T \Sigma_g^{-1}(Z_{i,g} - Z_{j,g})$     # Pairwise Mahalanobis distances
9.      $w_{ij}^{(g)} \leftarrow \frac{1}{1+D_{ij}^{(g)}}$     # Similarity weights
10.     $E^{(g)} \leftarrow KNN(D^{(g)})$     # Prune edges to k nearest neighbors
11.     $M_{ij}^{(g)} = 1\left(D_{ij}^{(g)} \leq \tau\right)$, $\widehat{W}^{(g)} = \widetilde{W}^{(g)} \odot M^{(g)}$     # Threshold-based edge pruning
12.    end for
13.   Build multi-relational graph $\mathcal{G} = \{V, E^{(g)}, W^{(g)}\}$;
14.   Apply relational GAT layers $\rightarrow$ obtain $H^{(1)}$, then $H^{(2)}$ through gated fusion;
15.   Compute class probabilities for query nodes $\hat{\mathcal{Y}} = Softmax\left(MLP\left(H_{query}^{(2)}\right)\right)$;
16.   Evaluate focal loss $\mathcal{L} = \mathcal{L}_{focal}(\hat{\mathcal{Y}}, \mathcal{Y})$;
17.   Update parameters $\theta \leftarrow \theta - \eta \nabla_\theta \mathcal{L}$;
18. end for
19. return $\theta$

---

## 4. Evaluation

This section evaluates the proposed MRC-GAT method for multimodal Alzheimer's disease classification with TADPOLE and NACC datasets. The model's performance is compared with state-of-the-art baselines on three and binary classification tasks. The subsequent sections discuss training configuration, datasets, evaluation metrics, results and comparisons, and finally provide an interpretability analysis.

### 4.1. Training Configuration

The meta-training takes place in the context of the episodic meta-learning paradigm, aimed at achieving inductive generalization to unseen subjects under a few-shot learning setting. Each episode consists of 10 support examples per class, and one query sample, while there are 32 episodes sampled in every iteration. Additionally, in each episode, the feature vectors are rank-Gaussianized, along with covariance shrinkage, along with the formation of directed KNN graphs, where the parameter values are set to $k = 6$, and a threshold $\tau = 1$ is applied to prune weak edges. The code for the experiment has been written in Python version 3.10, PyTorch version 2.2.0, and CUDA version 12.1, running inside the Google Colab platform, set up with an Nvidia Tesla T4 GPU, having a capacity of 16 GB GPU RAM.

The Relational GAT module has two attention layers having attention heads of size four and two, respectively, where node-wise gated fusion is applied afterwards. The parameter settings for the optimisation process include the use of the Adam optimiser, where the learning rate starts at 0.01, and a total of 1200 iterations is performed. The dropout strategy has a rate of 0.2, applied to the attention mechanisms. The proposed approach also follows a five-fold cross-validation process. Table 2 provides a summary of the training parameters.

Table 2. Summary of Training Configuration

| Parameter | Setting |
|---|---|
| method | Episodic meta-learning |
| Hardware | NVIDIA Tesla T4 (16 GB VRAM), 2 vCPUs, CUDA 12.1 |
| Software | Python 3.10, PyTorch 2.2.0 |
| Graph pruning | KNN (k = 6), τ = 1 |
| Relational GAT | 2 layers (4 heads for the first layer and 2 heads for the second layer) |
| Optimizer | Adam, learning rate = 0.01 |
| Training Iterations | 1200 |
| Batch Size | 32 |
| Regularization | Dropout = 0.2 |
| Validation Protocol | five-fold cross-validation |

## 4.2. Datasets

Two complementary multimodal datasets were used to evaluate the proposed model across distinct clinical and acquisition conditions. Together, they provide heterogeneous combinations of imaging, cognitive, and demographic features, enabling a comprehensive assessment of model performance and generalisation. The following subsections briefly describe the TADPOLE and NACC datasets.

**Tadpole Dataset:** For both training and evaluation of the model, the TADPOLE dataset [35] was used. This dataset was derived from the Alzheimer's Disease Neuroimaging Initiative (ADNI), which presents comprehensive multimodal clinical data, including demographic and genetic risk factors, cognitive assessment scores, and neuroimaging biomarkers. These complementary modalities together describe the multifaceted aspects of Alzheimer's disease and provide a strong basis for comparing models that incorporate heterogeneous feature domains under a single learning method. A single cross-sectional snapshot from each participant was used, emphasizing the joint representation of multimodal features rather than their temporal evolution. This setting reflects a clinically realistic diagnostic scenario, in which medical decisions are typically made based on baseline examinations. Furthermore, the dataset encompasses a broad spectrum of disease stages, ranging from cognitively normal to mild cognitive impairment and Alzheimer's disease, enabling a comprehensive evaluation of the model's generalization ability across heterogeneous patient profiles. The proposed MRC-GAT architecture is specifically designed to manage multimodal variability and relational complexity, which makes the TADPOLE dataset an appropriate and challenging benchmark to validate the effectiveness of the proposed approach.

**NACC Dataset:** The National Alzheimer's Coordinating Center (NACC) dataset [36] is a comprehensive, multimodal repository on Alzheimer's disease, combining diverse data types, such as clinical assessments, cognitive scores, neuroimaging, genetic markers, and biomarkers. Collected longitudinally from several Alzheimer's Disease Research Centers (ADRCs) across the United States, it covers a wide range of participants spanning cognitively normal to mild cognitive impairment and Alzheimer's disease stages. While this dataset does not represent the general population because of its specific focus on research cohorts, the standardized protocols assure that features are of high quality and well harmonized, which will permit robust modeling of disease progression and diagnostic patterns. This makes NACC especially suitable for assessing advanced machine learning methods, such as the proposed MRC-GAT, because it allows testing generalization across heterogeneous multimodal inputs under real-world clinical variability.

## 4.3. Evaluation Metrics

Performance of the proposed model is evaluated using a set of widely adopted classification metrics. This evaluation protocol allows clear and consistent comparison across episodes, with reflection of clinical significance regarding discrimination among the diagnostic categories CN, MCI, and AD. The metrics definitions are as follows:

- **Model Performance Evaluation Metric**

The performance of the proposed model in this study was evaluated using the accuracy metric. Accuracy measures the proportion of correctly predicted samples relative to the total number of samples and is defined as follows:

$$ACC = \frac{TP + TN}{TP + TN + FP + FN} \tag{21}$$

where TP, TN, FP, and FN denote true positive, true negative, false positive, and false negative counts, respectively

- **Area Under the Receiver Operating Characteristic Curve (AUC)**

To test the discriminative power of the classifier, the Area Under the ROC Curve is calculated as:

$$\int_0^1 \text{ROC}(f)\, df \tag{22}$$

where $ROC(f)$ is the Receiver Operating Characteristic curve obtained by varying the decision threshold $f$. The Micro-AUC is obtained by aggregating the true positive and false positive counts across all the diagnostic classes and subsequently calculating the area under the combined ROC curve, that is:

$$AUC_{micro} = AUC\left(\frac{\sum_{c=1}^{C} TP_c}{\sum_{c=1}^{C}(TP_c + FN_c)}, \frac{\sum_{c=1}^{C} FP_c}{\sum_{c=1}^{C}(FP_c + TN_c)}\right) \tag{23}$$

where $TP_c$, $FP_c$, $FN_c$, and $TN_c$ denote the true positive, false positive, false negative, and true negative counts for class $c$, respectively, and $C$ represents the total number of diagnostic classes.

- **DeepROC-Based Analysis**

In addition to the traditional metrics, a DeepROC [37] analysis was performed, quantifying model sensitivity across specific False Positive Rate (FPR) intervals. Building on the work of Carrington et al. [38], the FPR domain [0,1] was discretized into three sub-ranges, Group 1 [0,0.33], Group 2 [0.33,0.67], and Group 3 [0.67,1] to test the diagnostic stability at different clinical risk levels. For each interval, the average sensitivity ($\overline{se}_i$) and average specificity ($\overline{sp}_i$) were computed as:

$$\overline{se}_i = \frac{1}{\Delta x} \int_{x_1}^{x_2} r(x)\, dx \tag{24}$$

$$\overline{sp}_i = \frac{1}{\Delta y} \int_{y_1}^{y_2} [1 - r^{-1}(y)]\, dy \tag{25}$$

where $r(x)$ denotes the ROC function mapping the false-positive rate $x$ into the true-positive rate $y$, and $\Delta x = x_2 - x_1$, $\Delta y = y_2 - y_1$ represent the width of the integration intervals. Using these, the normalized interval-based $\text{AUC}_{ni}$ is then formulated as:

$$\text{AUC}_{ni} = \frac{\Delta x}{\Delta x + \Delta y}\overline{se}_i + \frac{\Delta y}{\Delta x + \Delta y}\overline{sp}_i \tag{26}$$

which integrates both average sensitivity and specificity to provide a localized assessment of model reliability under specific risk regions. In summary, these metrics jointly provide a comprehensive assessment of the model's diagnostic performance. The next subsection discusses the experimental results based on these evaluation measures.

### 4.4. Experimental Result

This section presents an extensive evaluation of the proposed MRC-GAT model on multiple experimental settings. A detailed performance investigation is conducted on the TADPOLE and NACC datasets for the three-class and binary classification tasks, including comparisons with state-of-the-art graph-based baselines. Also, interpretability analyses are given to illustrate how the model utilizes the multimodal information and relational structures during the course of making diagnostic process.

#### 4.4.1. Tadpole Results

To validate the proposed MRC-GAT model, experiments were first performed on the widely used Tadpole dataset, which offers multimodal information across MRI, cognitive, and demographic features, enabling a comprehensive evaluation of model generalization and cross-modal fusion. The overall effectiveness and training dynamics of the MRC-GAT model were investigated through various metrics, including training loss trends, ROC curves from cross-validation, and confusion matrices. These elements served to appraise the model's steadiness, ability to differentiate classes, and uniformity over different folds. As illustrated in Fig. 3(a), the training loss exhibits a rapid decline during the initial iterations, followed by gradual stabilization. This pattern demonstrates a smooth convergence process with low fluctuation, indicating that the proposed optimization strategy effectively maintains training stability. The relatively narrow standard deviation across folds further supports the reproducibility and robustness of the episodic meta-learning process introduced earlier. Fig. 3(b) presents the cross-fold micro-averaged ROC curves across validation folds. The model achieves consistently high performance, with AUC values ranging from 0.990 to 1.000 and a mean AUC of $0.997 \pm 0.004$. The tight alignment of these curves and minimal spread across partitions underscores the model's solid adaptability to varied data groupings. Besides, the confusion matrix shown in Fig. 3(c) provides a detailed view of the classification outcomes across diagnostic categories. The model accurately identifies CN and AD subjects with recognition rates above 99%, while MCI cases achieve an accuracy of 90.5%. Misclassifications are primarily observed between adjacent cognitive stages, CN-MCI or MCI-AD, which is consistent with the clinical continuity of Alzheimer's disease progression.

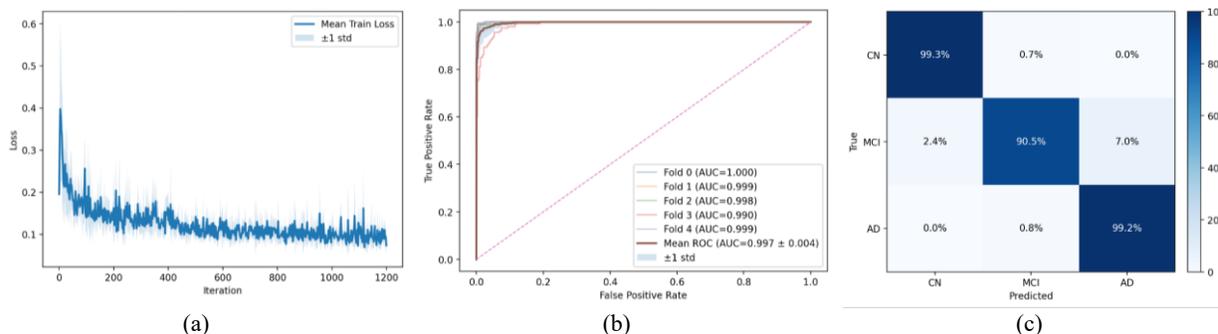

(a)                  (b)                  (c)

**Fig. 3.** Results on the TADPOLE dataset: (a) Training loss curve; (b) Cross-fold micro-averaged ROC curves; (c) Confusion matrix for CN, MCI, and AD classes.

- **Three-class Classification**

A comparative evaluation was conducted between the proposed MRC-GAT model and several representative graph-based baselines that have previously been applied to the Tadpole dataset. The selected baselines reflect the chronological development of multimodal graph strategies concerning Alzheimer's disease graph learning, including attention-based fusion and multigraph screening approaches. Specifically, AMGNN [18], EVGCN [39], MMAF [19], UPGT [40], and MGCS-GCN [20] were used as state-of-the-art references, all of which reported competitive performance on the same dataset, forming a robust benchmark for comparison. As can be seen from Table 3, the proposed MRC-GAT model outperforms all others in the three-class classification task, with an accuracy of 96.87±1.05% and an AUC of 99.59±1.16%, which outperforms the state-of-the-art graph-based and multimodal methods with a clear margin. The best performance among baselines is MGCS-GCN, whose accuracy and AUC are 94.10±2.32% and 97.90±1.12%, respectively. These results all point to the strong classification performance of the MRC-GAT model.

Table 3. Three-class classification experimental results on the Tadpole Dataset

| Author | Method | Published | Dataset | CN vs MCI vs AD | |
|---|---|---|---|---|---|
| | | | | ACC (%) | AUC (%) |
| Song X et al. [18] | AMGNN | 2021 | Tadpole | 94.06 ± 0.38 | ---- |
| Huang Y et al. [39] | EVGCN | 2022 | Tadpole | 87.51 ± 2.34 | 90.97 ± 2.15 |
| Yang F et al. [19] | MMAF | 2023 | Tadpole | 92.80 ± 0.92 | 93.32 ± 1.78 |
| Pellegrini C et al. [40] | UPGT | 2023 | Tadpole | 92.59 ± 3.64 | 96.96 ± 2.32 |
| Huabin W et al. [20] | MGCS-GCN | 2024 | Tadpole | 94.10 ± 2.32 | 97.90 ± 1.12 |
| **Proposed method** | | | **Tadpole** | **96.87 ± 1.05** | **99.70 ± 0.4** |

- **Binary Classification**

To further confirm the diagnostic effectiveness demonstrated in the multiclass experiments, three additional binary classification tasks were conducted on the Tadpole dataset, namely CN vs AD, MCI vs AD, and CN vs MCI. These pairwise settings reflect different stages of cognitive decline and provide a finer evaluation of the model's sensitivity and reliability under clinically relevant conditions. As summarized in Tables 4-6 and visualized in Fig. 4, the proposed MRC-GAT outperforms four representative baselines, Spectral-GCN [41], Inception-GCN [42], MMAF [19], and MGCS-GCN[20] across all evaluation metrics, including AUC, $AUC_{ni}$, and the DeepROC-derived average sensitivity $\overline{se}_i$ and specificity $\overline{sp}_i$. These results extend the stability trends observed in the previous section. Among the three experiments, MCI vs AD remains the most challenging because of the significant clinical and imaging overlap between mild impairment and early dementia. As reported in Table 4, MRC-GAT attains the best overall performance with an $AUC_{ni}$ of 98.8 %, exceeding MGCS-GCN (98.4 %), MMAF (96.1 %), and Inception-GCN (93.8 %). As illustrated in Fig. 5(a), the DeepROC analysis shows an average sensitivity of 96.5% and specificity of 98.6% within Group 1, and the low-FPR interval is crucial for clinical safety. The corresponding ROC curve rises sharply near the origin, reflecting reliable detection even under minimal false-positive conditions.

Table 4. DeepRoc analysis in the MCI versus AD experiment on the Tadpole dataset

| FPR | [0,1] | [0,0.33] | [0.33, 0.67] | [0.67, 1] |
|---|---|---|---|---|
| Pres. Risk | All | Group1 | Group2 | Group2 |
| $AUC_{n_i}$ | 92.4% | | Spectral-GCN | |
| AUC | 92.4% | 92.7% | 89.3% | 94.8% |
| $\overline{se}_i$ | 92.4% | 81.9% | 96.2% | 99.0% |
| $\overline{sp}_i$ | 92.4% | 96.6% | 57.1% | 15.4% |
| $AUC_{n_i}$ | 93.8% | | Inception-GCN | |
| AUC | 93.8% | 92.0% | 94.3% | 100% |
| $\overline{se}_i$ | 93.8% | 82.2% | 99.1% | 100% |
| $\overline{sp}_i$ | 93.8% | 95.4% | 60.3% | 0% |
| $AUC_{n_i}$ | 96.1% | | MMAF | |
| AUC | 96.1% | 94.1% | 100% | 100% |
| $\overline{se}_i$ |  | 96.1% | 88.2% | 100% | 100% |
| $\overline{sp}_i$ |  | 96.1% | 96.1% | 0% | 0% |
| $AUC_{n_i}$ | 98.4% | | MGCS-GCN | | |
| AUC | 98.4% | 97.6% | 100% | 100% |
| $\overline{se}_i$ | 98.4% | 95.1% | 100% | 100% |
| $\overline{sp}_i$ | 98.4% | 98.4% | 0% | 0% |
| $AUC_{n_i}$ | 98.8% | | **Proposed method** | | |
| AUC | 98.8% | 98.4% | 100% | 100% |
| $\overline{se}_i$ | 98.8% | 96.5% | 100% | 100% |
| $\overline{sp}_i$ | 98.8% | 98.6% | 0% | 0% |

As shown in Table 5, for the CN vs AD experiment, where the phenotypic gap between normal and diseased groups is broader, MRC-GAT achieves an $AUC_{ni}$ of 99.8 %, outperforming MGCS-GCN (99.6 %), MMAF (99.2 %), and Inception-GCN (98.4 %). The ROC curve in Fig. 5(b) demonstrates a nearly vertical ascent at the beginning of the axis, signifying extremely high sensitivity, 99.5% coupled with equally high specificity, 99.7% in the early detection zone.

Table 5. DeepRoc analysis in the CN versus AD experiment on the Tadpole dataset

| FPR | [0,1] | [0,0.33] | [0.33, 0.67] | [0.67, 1] |
|---|---|---|---|---|
| Pres. Risk | All | Group1 | Group2 | Group2 |
| $AUC_{n_i}$ | 97.1% | | Spectral-GCN | |
| AUC | 97.1% | 98.2% | 91.0% | 100% |
| $\overline{se}_i$ | 97.1% | 93.3% | 97.9% | 100% |
| $\overline{sp}_i$ | 97.1% | 100% | 56.2% | 0% |
| $AUC_{n_i}$ | 98.4% | | Inception-GCN | |
| AUC | 98.4% | 97.5% | 100% | 100% |
| $\overline{se}_i$ | 98.4% | 95.1% | 100% | 100% |
| $\overline{sp}_i$ | 98.4% | 98.4% | 0% | 0% |
| $AUC_{n_i}$ | 99.2% | | MMAF | |
| AUC | 99.2% | 98.8% | 100% | 100% |
| $\overline{se}_i$ | 99.2% | 97.5% | 100% | 100% |
| $\overline{sp}_i$ | 99.2% | 99.2% | 0% | 0% |
| $AUC_{n_i}$ | 99.6% | | MGCS-GCN | |
| AUC | 99.6% | 99.6% | 100% | 100% |
| $\overline{se}_i$ | 99.6% | 99.1% | 100% | 100% |
| $\overline{sp}_i$ | 99.6% | 99.7% | 0% | 0% |
| $AUC_{n_i}$ | 99.8% | | **Proposed method** | |
| AUC | 99.8% | 99.7% | 100% | 100% |
| $\overline{se}_i$ | 99.8% | 99.5% | 100% | 100% |
| $\overline{sp}_i$ | 99.8% | 99.7% | 0% | 0% |

In the CN vs MCI comparison, as shown in Table 6, which evaluates the earliest transition stage of cognitive decline, the proposed model again achieves the highest performance, reaching $AUC_{ni}$ = 99.9%, surpassing MGCS-GCN (99.7%), MMAF (97.2%), and Inception-GCN (96.3%). As illustrated in Fig. 5(c), the ROC trajectory of the proposed model demonstrates a steep initial ascent and near-saturation behavior, indicating outstanding detection capability in the clinically critical low-FPR region (Group 1). Within this region, the model attains an average sensitivity of 99.1% and specificity of 99.7%.

**Table 6.** DeepRoc analysis in the CN versus MCI experiment on the Tadpole dataset

| FPR<br>Pres. Risk | [0,1]<br>All | [0,0.33]<br>Group1 | [0.33, 0.67]<br>Group2 | [0.67, 1]<br>Group۳ |
|---|---|---|---|---|
| $AUC_{n_i}$ | 92.8% | | Spectral-GCN | |
| AUC | 92.8% | 94.8% | 82.5% | 100% |
| $\overline{se}_i$ | 92.8% | 83.7% | 94.6% | 100% |
| $\overline{sp}_i$ | 92.8% | 99.2% | 54.2% | 0% |
| $AUC_{n_i}$ | 96.3% | | Inception-GCN | |
| AUC | 96.3% | 94.4% | 100% | 100% |
| $\overline{se}_i$ | 96.3% | 88.8% | 100% | 100% |
| $\overline{sp}_i$ | 96.3% | 96.3% | 0% | 0% |
| $AUC_{n_i}$ | 97.2% | | MMAF | |
| AUC | 97.2% | 97.8% | 93.7% | 98.3% |
| $\overline{se}_i$ | 97.2% | 93.8% | 97.7% | 99.9% |
| $\overline{sp}_i$ | 97.2% | 99.2% | 50.4% | 28.3% |
| $AUC_{n_i}$ | 99.7% | | MGCS-GCN | |
| AUC | 99.7% | 99.5% | 100% | 100% |
| $\overline{se}_i$ | 99.7% | 98.9% | 100% | 100% |
| $\overline{sp}_i$ | 99.7% | 99.6% | 0% | 0% |
| $AUC_{n_i}$ | 99.9% | | **Proposed method** | |
| AUC | 99.9% | 99.6% | 100% | 100% |
| $\overline{se}_i$ | 99.9% | 99.1% | 100% | 100% |
| $\overline{sp}_i$ | 99.9% | 99.7% | 0% | 0% |

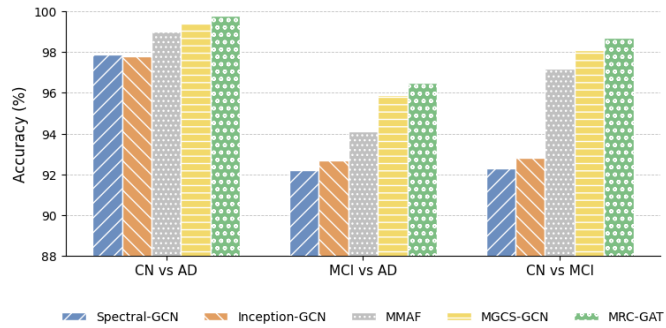

**Fig. 4.** Binary classification experimental results of different models on the Tadpole dataset.

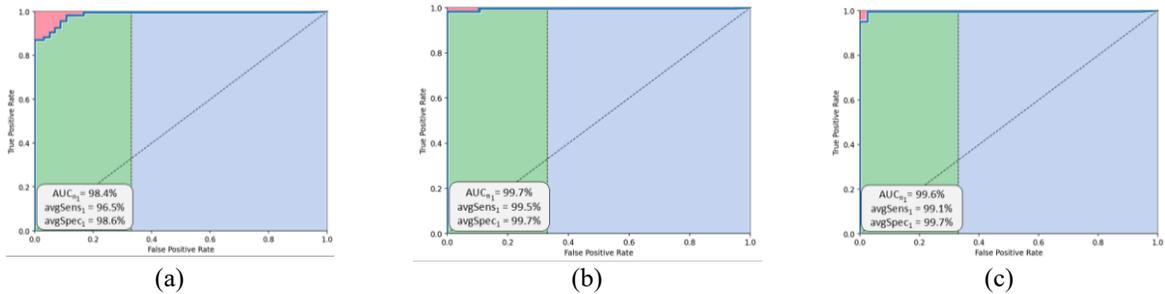

**Fig. 5.** DeepROC comparison visualization on the Tadpole dataset. (a) MCI&AD; (b) CN&AD; (c) CN&MCI

These binary classification results, along with the previous multiclass evaluation, altogether demonstrated that MRC-GAT achieved not only the highest overall diagnostic accuracy but also maintained excellent sensitivity and specificity under clinically relevant conditions. This consistent performance confirms the strong diagnostic capability and stability of the proposed model on the TADPOLE dataset, highlighting its effectiveness in integrating and interpreting complex multimodal relationships within a representative Alzheimer's cohort and naturally supporting its extension to the NACC dataset.

### 4.4.2. NACC Results

To further assess the model's robustness under real-world variability, the proposed MRC-GAT was next evaluated on the independent NACC dataset, which contains heterogeneous clinical and imaging data collected across multiple centers. As shown in Fig. 6(a), the training loss continuously reduces over the training iterations. Though there is more fluctuation at the beginning because of the higher variability of the NACC dataset but the curve stabilizes over time. Fig. 6(b) presents the micro-averaged ROC curves across the five validation folds of the NACC dataset. The curves show a consistently strong separation between classes, with AUC scores ranging from 0.955 to 0.997 and a mean value of $0.980 \pm 0.015$. The limited spread observed across folds suggests that the model performs reliably despite the greater variability inherent in NACC's multimodal data. In addition, Fig. 6(c) displays the row-normalized confusion matrix, summarizing the distribution of predictions across diagnostic groups. The model achieves a notably high detection rate for AD (97.9%), while MCI and CN cases reach 87.7% and 85.6%, respectively. The majority of errors arise between neighboring cognitive stages, particularly CN-MCI, reflecting the subtle and gradual transitions characteristic of early Alzheimer's progression.

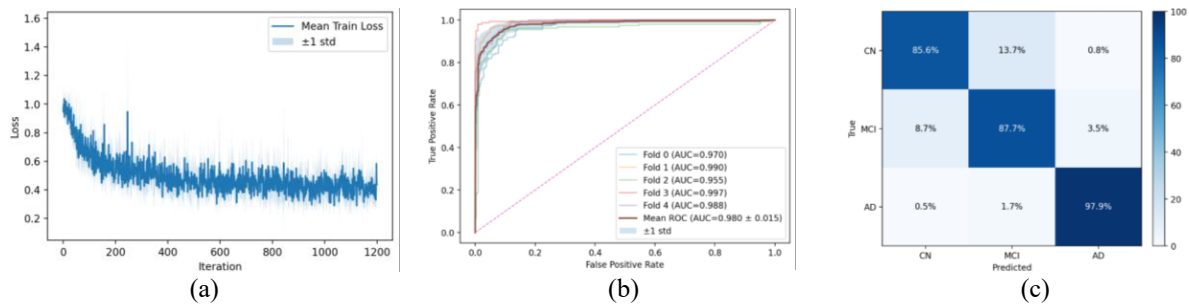

(a) (b) (c)

**Fig. 6.** Results on the NACC dataset: (a) Training loss curve; (b) Cross-fold micro-averaged ROC curves; (c) Confusion matrix for CN, MCI, and AD classes.

Overall, these details illustrate that MRC-GAT upholds steady predictive trustworthiness and separation ability across datasets with differing collection methods.

- **Three-class classification**

To more thoroughly assess the generalisation capability of the MRC-GAT model under heterogeneous clinical conditions, a three-class evaluation was also conducted on the NACC dataset. Compared with TADPOLE, NACC exhibits higher variability and a less balanced class distribution, providing a more challenging benchmark for examining the model's robustness and adaptability to real-world diagnostic settings. As summarized in Table 7, the proposed MRC-GAT achieves the highest diagnostic performance among all evaluated methods, attaining $92.31 \pm 1.02$ % accuracy and $98.0 \pm 1.5$ % AUC. These results surpass several well-known baselines, including Pop-GCN [41], Inception-GCN [42], EV-GCN [39], and MGCS-GCN [20], by a clear margin. Among the competing approaches, MGCS-GCN demonstrates the best baseline results with $90.21 \pm 2.89$ % accuracy and $92.13 \pm 1.02$ % AUC. however, MRC-GAT further improves both metrics, confirming its superior capability to model complex cross-modal dependencies and maintain diagnostic stability across heterogeneous clinical cohorts. In comparison with the findings from the previous section, this outcome highlights the model's consistent behavior and effective generalisation under varying dataset characteristics.

**Table 7.** Three-class classification experimental results on the NACC Dataset

| Author | Method | Published | Dataset | CN vs MCI vs AD | |
|---|---|---|---|---|---|
| | | | | ACC (%) | AUC (%) |
| Parisot S et al. [41] | popGCN | 2017 | NACC | 80.11 ± 3.64 | 81.36 ± 2.85 |
| Kazi A et al. [42] | Inception-GCN | 2019 | NACC | 79.26 ± 1.98 | 80.34 ± 1.25 |
| Huang Y et al. [39] | EV-GCN | 2022 | NACC | 86.35 ± 2.35 | 84.73 ± 1.82 |
| wang W et al. [20] | MGCS-GCN | 2024 | NACC | 90.21 ± 2.89 | 92.13 ± 1.02 |
| **Proposed method** | | | **NACC** | **92.31 ± 1.02** | **98.0 ± 1.5** |

- **Binary Classification**

Corresponding to the patterns presented in the TADPOLE dataset, the binary studies undertaken on the NACC cohort also depict the superiority of MRC-GAT in more heterogeneous clinical conditions. The model continues to provide leading accuracy, sensitivity, and specificity throughout the different tasks compared to all other tested models, as summarized in Tables 8-10 and illustrated in Figs 7,8. The proposed MRC-GAT consistently outperforms the baseline models, Spectral-GCN [41], Inception-GCN [42], MMAF [19], and MGCS-GCN [20] across all binary classification tasks. As shown in Table 8, in the MCI vs AD experiment, where disease stages exhibit significant overlap, MRC-GAT achieves the highest performance with an $AUC_{ni}$ of 97.1%, exceeding MGCS-GCN (95.8%) and MMAF (91.8%) by a considerable margin. Within the clinically crucial low-FPR region (Group 1), the model achieves an average sensitivity of 94.3% and specificity of 96.9%, reflecting strong discriminability under challenging diagnostic boundaries. The ROC curve in Fig. 8 (a) confirms a steep initial rise and near-saturation trend, indicating accurate classification with minimal false-positive rates. These findings align with the cross-dataset consistency observed earlier and verify the model's capacity to differentiate subtle cognitive transitions effectively.

As reported in Table 9, the CN vs AD experiment, which involves a more distinct phenotypic separation, the proposed model achieves an $AUC_{ni}$ of 99.2%, surpassing MGCS-GCN (98.9%), MMAF (97.2%), and Inception-GCN (93.9%). The DeepROC analysis further demonstrates an average sensitivity of 97.1% and specificity of 99.1% in Group 1, emphasizing that the model retains high diagnostic precision even in early detection regions. As illustrated in Fig. 8(b), the ROC trajectory of MRC-GAT rises sharply near the origin, indicating reliable detection of AD subjects while maintaining an exceptionally low false-positive rate. For the CN vs MCI task, which represents the earliest and most subtle stage of cognitive decline, MRC-GAT once again achieves the best overall results with an $AUC_{ni}$ of 97.5%, outperforming MGCS-GCN (96.7%), MMAF (94.8%), and Inception-GCN (91.6%). The DeepROC evaluation in Table 10 shows that within Group 1, the model achieves an average sensitivity of 91.1% and specificity of 98.2%, surpassing all baseline methods. The ROC visualization in Fig. 8 (c) demonstrates a steep ascent with near-saturation behavior, indicating precise recognition in early diagnostic zones and highlighting the model's strength in identifying mild cognitive impairment.

Finally, these binary classification outcomes complement the three-class experiments presented earlier, confirming that MRC-GAT not only achieves state-of-the-art accuracy but also maintains balanced sensitivity and specificity across diverse diagnostic settings. This consistency across datasets underscores the model's potential as a reliable and clinically applicable tool for multimodal Alzheimer's disease diagnosis.

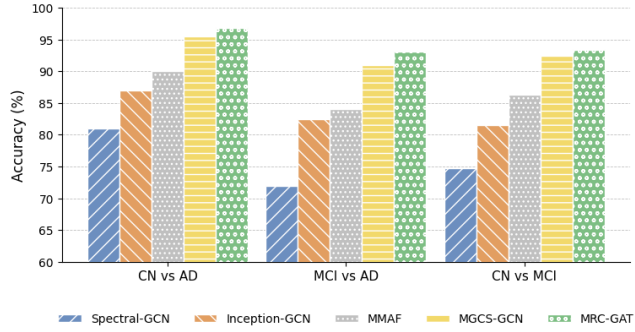

**Fig. 7**. Binary classification experimental results of different models on the NACC dataset.

**Table 8.** DeepRoc analysis in the MCI versus AD experiment on the NACC dataset

| FPR<br>Pres. Risk | [0,1]<br>All | [0,0.33]<br>Group1 | [0.33, 0.67]<br>Group2 | [0.67, 1]<br>Group۳ |
|---|---|---|---|---|
| $AUC_{n_i}$ | 83.7% | | Spectral-GCN | |
| AUC | 83.7% | 80.6% | 89.0% | 89.1% |
| $\overline{se}_i$ | 83.7% | 59.9% | 94.0% | 97.4% |
| $\overline{sp}_i$ | 83.7% | 88.1% | 40.6% | 7.8% |
| $AUC_{n_i}$ | 91.6% | | Inception-GCN | |
| AUC | 91.6% | 89.1% | 93.8% | 100% |
| $\overline{se}_i$ | 91.6% | 77.0% | 98.3% | 100% |
| $\overline{sp}_i$ | 91.6% | 93.2% | 50.0% | 0% |
| $AUC_{n_i}$ | 91.8% | | MMAF | |
| AUC | 91.8% | 90.2% | 94.8% | 94.6% |
| $\overline{se}_i$ | 91.8% | 79.2% | 97.3% | 98.8% |
| $\overline{sp}_i$ | 91.8% | 94.0% | 45.2% | 9.7% |
| $AUC_{n_i}$ | 95.8% | | MGCS-GCN | |
| AUC | 95.8% | 94.3% | 97.3% | 100% |
| $\overline{se}_i$ | 95.8% | 87.8% | 99.4% | 100% |
| $\overline{sp}_i$ | 95.8% | 96.4% | 54.8% | 0% |
| $AUC_{n_i}$ | 97.1% | | **Proposed method** | |
| AUC | 97.1% | 95.4% | 98.1% | 100% |
| $\overline{se}_i$ | 97.1% | 94.3% | 99.7% | 100% |
| $\overline{sp}_i$ | 97.1% | 96.9% | 58.9% | 0% |

**Table 9.** DeepRoc analysis in the CN versus AD experiment on the NACC dataset

| FPR<br>Pres. Risk | [0,1]<br>All | [0,0.33]<br>Group1 | [0.33, 0.67]<br>Group2 | [0.67, 1]<br>Group۳ |
|---|---|---|---|---|
| $AUC_{n_i}$ | 91.0% | | Spectral-GCN | |
| AUC | 91.0% | 87.8% | 93.6% | 100% |
| $\overline{se}_i$ | 91.0% | 73.6% | 99.4% | 100% |
| $\overline{sp}_i$ | 91.0% | 92.9% | 63.5% | 0% |
| $AUC_{n_i}$ | 93.9% | | Inception-GCN | |
| AUC | 93.9% | 93.7% | 96.9% | 92.1% |
| $\overline{se}_i$ | 93.9% | 86.2% | 96.9% | 98.8% |
| $\overline{sp}_i$ | 93.9% | 96.3% | 0% | 20.6% |
| $AUC_{n_i}$ | 97.2% | | MMAF | |
| AUC | 97.2% | 95.8% | 100% | 100% |
| $\overline{se}_i$ | 97.2% | 91.6% | 100% | 100% |
| $\overline{sp}_i$ | 97.2% | 97.2% | 0% | 0% |
| $AUC_{n_i}$ | 98.9% | | MGCS-GCN | |
| AUC | 98.9% | 98.3% | 100% | 100% |
| $\overline{se}_i$ | 98.9% | 96.6% | 100% | 100% |
| $\overline{sp}_i$ | 98.9% | 98.9% | 0% | 0% |
| $AUC_{n_i}$ | 99.2% | | **Proposed method** | |
| AUC | 99.2% | 98.9% | 100% | 100% |
| $\overline{se}_i$ | 99.2% | 97.1% | 100% | 100% |
| $\overline{sp}_i$ | 99.2% | 99.1% | 0% | 0% |

**Table 10.** DeepRoc analysis in the CN versus MCI experiment on the NACC dataset

| FPR | [0,1] | [0,0.33] | [0.33, 0.67] | [0.67, 1] |
|---|---|---|---|---|
| Pres. Risk | All | Group1 | Group2 | Group3 |
| $AUC_{n_i}$ | 82.7% | | Spectral-GCN | |
| AUC | 82.7% | 84.4% | 80.5% | 80.5% |
| $\overline{se}_i$ | 82.7% | 64.4% | 87.8% | 96.2% |
| $\overline{sp}_i$ | 82.7% | 92.5% | 51.9% | 18.5% |
| $AUC_{n_i}$ | 91.6% | | Inception-GCN | |
| AUC | 91.6% | 92.0% | 83.9% | 100% |
| $\overline{se}_i$ | 91.6% | 80.7% | 94.2% | 100% |
| $\overline{sp}_i$ | 91.6% | 96.1% | 44.2% | 0% |
| $AUC_{n_i}$ | 94.8% | | MMAF | |
| AUC | 94.8% | 93.6% | 94.3% | 100% |
| $\overline{se}_i$ | 94.8% | 85.2% | 99.3% | 100% |
| $\overline{sp}_i$ | 94.8% | 96.6% | 62.3% | 0% |
| $AUC_{n_i}$ | 96.7% | | MGCS-GCN | |
| AUC | 96.7% | 96.1% | 96.2% | 100% |
| $\overline{se}_i$ | 96.7% | 90.6% | 99.6% | 100% |
| $\overline{sp}_i$ | 96.7% | 97.9% | 62.5% | 0% |
| $AUC_{n_i}$ | 97.5% | | **Proposed method** | |
| AUC | 97.5% | 97.3% | 97.1% | 100% |
| $\overline{se}_i$ | 97.5% | 91.1% | 99.7% | 100% |
| $\overline{sp}_i$ | 97.5% | 98.2% | 65.4% | 0% |

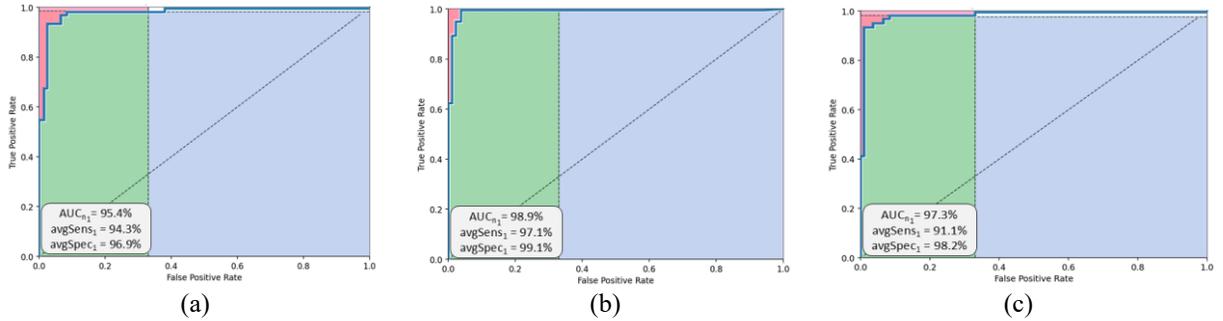

(a)     (b)     (c)

**Fig. 8.** DeepROC comparison visualization on the NACC dataset. (a) MCI&AD; (b) CN&AD; (c) CN&MCI

## 4.5. Interpretability Analysis

The interpretability of the proposed MRC-GAT model was investigated to provide insights into how the model allocates attention across modalities and inter-subject relations. This was designed at two complementary levels of modality-level gating and edge-level attention, allowing for both global and localized perspectives on how multimodal features interact during episodic learning. Together, these two analyses illustrate how MRC-GAT achieves transparency and context-aware representation learning through its adaptive fusion mechanism. At the modality level, the gating heatmap across ten representative episodes, as depicted in Fig. 9, demonstrated that the relative importance of RF, COG, and MRI modalities is not fixed but dynamically changes. This implies that the model attentively adjusts the reliance on different feature sources with respect to changes in data composition and diagnostic context. For instance, in those episodes with significant cognitive variability, the COG modality was more activated, while MRI features were more prominent when structural biomarkers were more discriminative. Such fluctuations in the activations validate that the network effectively balances complementary information from heterogeneous modalities and avoids overdependence on any single feature domain. Such adaptive modulation at the modality level provides a global understanding of how the network manages multimodal interactions, thus laying the conceptual foundation for further edge-level interpretability analysis.

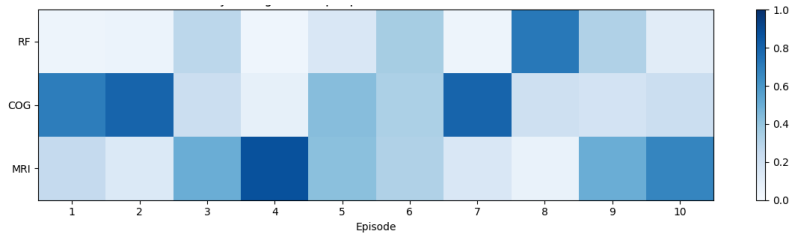

**Fig. 9.** Modality gating heatmap across ten episodes.

Based on the above global perspective, the edge-level interpretability result provides additional insight into how MRC-GAT captures the fine-grained relational dependencies between the subjects. as shown in Fig.10, there are three attention graphs that are modality-aware, where the thickness of the edges represents the attention weights. The result shows an asymmetric connectivity pattern where only some of the neighboring subjects strongly influence the target node representation. This observation indicates that the model successfully captures the important relational dependencies between the subjects in the clinical domain while suppressing the less informative ones. In addition, the variation in the importance of the edges associated with each modality shows that the influence of the relational dependencies depends on the modality, which captures the way in which similarities in demographics, cognition, and imaging influence the prediction representation. The consistency in the above result with the previously identified modality-level gating behavior provides evidence that the model provides consistent interpretability insights at hierarchical levels of reasoning.

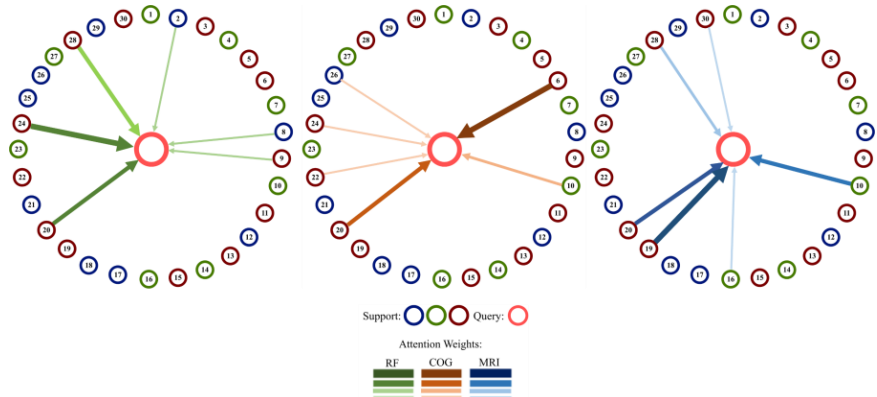

**Fig. 10.** Modality-specific attention graphs for RF, COG, and MRI.

Taken together, the results presented in Figs. 9 and 10 demonstrate that MRC-GAT achieves adaptive, context-dependent interpretability both globally and locally. The complementary findings between modality-level gating and edge-level attention analyses reinforce the transparency and consistency of the proposed model's decision process, closely aligning with the interpretability objectives established in the earlier methodology.

## 5. Conclusion and Future Work

This work introduced MRC-GAT as an adaptive and interpretable method for diagnosing Alzheimer's disease using multiple modalities. Integrating copula-based similarity alignment, relational attention modeling, and node-wise gated fusion within an episodic meta-learning paradigm effectively handled some serious limitations of traditional graph approaches, which are usually static and transductive. The proposed design allows for flexible handling of heterogeneous data sources spanning demographic, cognitive, and neuroimaging modalities while preserving both statistical consistency and generalizability. Extensive experimentation with the TADPOLE and NACC datasets yielded a state-of-the-art diagnostic accuracy of 96.87% and 92.31%, respectively, in three-class classification tasks. Furthermore, its consistent sensitivity and specificity across different binary evaluations confirmed the robustness and diagnostic reliability of the method in various clinical conditions. Moreover, the interpretability analyses both at modality and edge levels showed that this method dynamically adjusts the contribution of heterogeneous modalities and adaptively identifies clinically meaningful subject relationships. This transparency is important for a deep understanding of the model's internal decision process, enhancing trust and applicability in computer-aided medical diagnosis. While these results are promising, the current implementation bears a relatively high computational cost and involves parameter tuning. Future extensions are needed to improve computational efficiency, extend the method to longitudinal and predictive modeling of disease progression, and consider privacy-preserving and federated strategies for deployment in multi-center clinical environments.

**Ethics Statement**

Ethical approval and informed consent were not required for this study, as all data were obtained from publicly available, fully anonymized datasets (TADPOLE and NACC).

**Declaration of generative AI and AI-assisted technologies in the manuscript preparation process**

During the preparation of this work, the authors used ChatGPT (OpenAI) to assist in improving the fluency, clarity, and readability of the manuscript. After using this tool, the authors carefully reviewed, edited, and validated the content and take full responsibility for the integrity and accuracy of the published work.

**Competing Interests**

The authors declare that they have no competing interests.

## References

1. Kamal, M.S., et al., *Alzheimer's patient analysis using image and gene expression data and explainable-AI to present associated genes.* IEEE Transactions on Instrumentation and Measurement, 2021. 70: p. 1-7.
2. Nichols, E., et al., *Estimation of the global prevalence of dementia in 2019 and forecasted prevalence in 2050: an analysis for the Global Burden of Disease Study 2019.* The Lancet Public Health, 2022. 7(2): p. e105-e125.
3. Weller, J. and A. Budson, *Current understanding of Alzheimer's disease diagnosis and treatment.* F1000Research, 2018. 7: p. F1000 Faculty Rev-1161.
4. Adarsh, V., et al., *Multimodal classification of Alzheimer's disease and mild cognitive impairment using custom MKSCDDL kernel over CNN with transparent decision-making for explainable diagnosis.* Scientific Reports, 2024. 14(1): p. 1774.
5. Schmand, B., et al., *Value of neuropsychological tests, neuroimaging, and biomarkers for diagnosing Alzheimer's disease in younger and older age cohorts.* Journal of the American Geriatrics Society, 2011. 59(9): p. 1705-1710.

6. Citron, M., *Alzheimer's disease: treatments in discovery and development.* Nature neuroscience, 2002. 5(Suppl 11): p. 1055-1057.
7. Hao, X., et al., *Multi-modal neuroimaging feature selection with consistent metric constraint for diagnosis of Alzheimer's disease.* Medical image analysis, 2020. 60: p. 101625.
8. Tong, T., et al., *Multi-modal classification of Alzheimer's disease using nonlinear graph fusion.* Pattern recognition, 2017. 63: p. 171-181.
9. Farooq, A., et al. *A deep CNN based multi-class classification of Alzheimer's disease using MRI.* in *2017 IEEE International Conference on Imaging systems and techniques (IST).* 2017. IEEE.
10. Wu, Y., et al., *An attention-based 3D CNN with multi-scale integration block for Alzheimer's disease classification.* IEEE Journal of Biomedical and Health Informatics, 2022. 26(11): p. 5665-5673.
11. Li, J., et al., *3-D CNN-based multichannel contrastive learning for Alzheimer's disease automatic diagnosis.* IEEE Transactions on Instrumentation and Measurement, 2022. 71: p. 1-11.
12. Dwivedi, S., et al., *Multimodal fusion-based deep learning network for effective diagnosis of Alzheimer's disease.* IEEE MultiMedia, 2022. 29(2): p. 45-55.
13. Cheng, J., et al., *Alzheimer's disease prediction algorithm based on de-correlation constraint and multi-modal feature interaction.* Computers in biology and medicine, 2024. 170: p. 108000.
14. Ramana, T. and S. Nandhagopal, *Alzheimer disease detection and classification on magnetic resonance imaging (MRI) brain images using improved expectation maximization (IEM) and convolutional neural network (CNN).* Turkish Journal of Computer and Mathematics Education, 2021. 12(11): p. 5998-6006.
15. Jiang, W., et al., *CNNG: a convolutional neural networks with gated recurrent units for autism spectrum disorder classification.* Frontiers in Aging Neuroscience, 2022. 14: p. 948704.
16. Liu, S., et al., *Attention deficit/hyperactivity disorder classification based on deep spatio-temporal features of functional magnetic resonance imaging.* Biomedical Signal Processing and Control, 2022. 71: p. 103239.
17. Pfeifer, B., A. Saranti, and A. Holzinger, *GNN-SubNet: disease subnetwork detection with explainable graph neural networks.* Bioinformatics, 2022. 38(Supplement_2): p. ii120-ii126.
18. Song, X., M. Mao, and X. Qian, *Auto-metric graph neural network based on a meta-learning strategy for the diagnosis of Alzheimer's disease.* IEEE Journal of Biomedical and Health Informatics, 2021. 25(8): p. 3141-3152.
19. Yang, F., et al., *Multi-model adaptive fusion-based graph network for Alzheimer's disease prediction.* Computers in Biology and Medicine, 2023. 153: p. 106518.
20. Wang, H., et al., *A Multi-graph Combination Screening Strategy Enabled Graph Convolutional Network for Alzheimer's Disease Diagnosis.* IEEE Transactions on Instrumentation and Measurement, 2024.
21. Liu, L., et al., *Cascaded multi-modal mixing transformers for alzheimer's disease classification with incomplete data.* NeuroImage, 2023. 277: p. 120267.
22. Zhu, Q., et al., *Deep multi-modal discriminative and interpretability network for Alzheimer's disease diagnosis.* IEEE Transactions on Medical Imaging, 2022. 42(5): p. 1472-1483.
23. Wang, Y., et al., *Predicting long-term progression of Alzheimer's disease using a multimodal deep learning model incorporating interaction effects.* Journal of Translational Medicine, 2024. 22(1): p. 265.
24. Xi, Y., et al., *Predicting conversion of Alzheimer's disease based on multi-modal fusion of neuroimaging and genetic data.* Complex & Intelligent Systems, 2025. 11(1): p. 58.
25. Almohimeed, A., et al., *Explainable artificial intelligence of multi-level stacking ensemble for detection of Alzheimer's disease based on particle swarm optimization and the sub-scores of cognitive biomarkers.* Ieee Access, 2023. 11: p. 123173-123193.
26. Chen, J., et al., *Multimodal mixing convolutional neural network and transformer for Alzheimer's disease recognition.* Expert Systems with Applications, 2025. 259: p. 125321.

27. Qu, Z., et al., *A graph convolutional network based on univariate neurodegeneration biomarker for Alzheimer's disease diagnosis.* IEEE Journal of Translational Engineering in Health and Medicine, 2023. 11: p. 405-416.
28. Zhang, M., et al., *A feature-aware multimodal framework with auto-fusion for Alzheimer's disease diagnosis.* Computers in Biology and Medicine, 2024. 178: p. 108740.
29. Li, S. and R. Zhang, *A novel interactive deep cascade spectral graph convolutional network with multi-relational graphs for disease prediction.* Neural Networks, 2024. 175: p. 106285.
30. Valoor, A. and G. Gangadharan, *Unveiling the decision making process in Alzheimer's disease diagnosis: A case-based counterfactual methodology for explainable deep learning.* Journal of Neuroscience Methods, 2025. 413: p. 110318.
31. Bootun, D., et al., *ADAMAEX—Alzheimer's disease classification via attention-enhanced autoencoders and XAI.* Egyptian Informatics Journal, 2025. 30: p. 100688.
32. Jahan, S., et al., *Federated Explainable AI-based Alzheimer's Disease Prediction With Multimodal Data.* IEEE Access, 2025.
33. Rusch, T.K., M.M. Bronstein, and S. Mishra, *A survey on oversmoothing in graph neural networks.* arXiv preprint arXiv:2303.10993, 2023.
34. Wu, X., et al., *Demystifying oversmoothing in attention-based graph neural networks.* Advances in Neural Information Processing Systems, 2023. 36: p. 35084-35106.
35. Marinescu, R.V., et al., *TADPOLE challenge: prediction of longitudinal evolution in Alzheimer's disease.* arXiv preprint arXiv:1805.03909, 2018.
36. Beekly, D.L., et al., *The National Alzheimer's Coordinating Center (NACC) database: the uniform data set.* Alzheimer Disease & Associated Disorders, 2007. 21(3): p. 249-258.
37. Carrington, A.M., et al., *Deep ROC analysis and AUC as balanced average accuracy to improve model selection, understanding and interpretation.* arXiv preprint arXiv:2103.11357, 2021.
38. Carrington, A.M., et al., *Deep ROC analysis and AUC as balanced average accuracy, for improved classifier selection, audit and explanation.* IEEE Transactions on Pattern Analysis and Machine Intelligence, 2022. 45(1): p. 329-341.
39. Huang, Y. and A.C. Chung, *Disease prediction with edge-variational graph convolutional networks.* Medical Image Analysis, 2022. 77: p. 102375.
40. Pellegrini, C., N. Navab, and A. Kazi, *Unsupervised pre-training of graph transformers on patient population graphs.* Medical Image Analysis, 2023. 89: p. 102895.
41. Parisot, S., et al. *Spectral graph convolutions for population-based disease prediction*. in *International conference on medical image computing and computer-assisted intervention*. 2017. Springer.
42. Kazi, A., et al. *InceptionGCN: receptive field aware graph convolutional network for disease prediction*. in *International conference on information processing in medical imaging*. 2019. Springer.